\pdfoutput=1

\documentclass[11pt]{article}

\usepackage[]{emnlp2021}

\usepackage{times}
\usepackage{latexsym}

\usepackage[T1]{fontenc}

\usepackage[utf8]{inputenc}

\usepackage{microtype}

\usepackage{tabularx}
\usepackage{booktabs}
\usepackage{fontawesome5}
\usepackage{graphicx}
\usepackage{multirow}
\usepackage{tikz-dependency}
\usepackage{subcaption}
\usepackage{xcolor}
\usepackage{tabularx}
\usepackage{booktabs}
\usepackage{caption}
\usepackage{multirow}
\usepackage{makecell}
\usepackage{nicefrac}
\usepackage{xfrac}
\usepackage{amssymb}
\usepackage[flushleft]{threeparttable}
\usepackage{soul}
\usepackage{tablefootnote}

\newcommand{\projname}{\textsc{CrossRE}}

\title{CrossRE: A Cross-Domain Dataset for Relation Extraction}

\author{Elisa Bassignana\textsuperscript{\faCompass} \and Barbara Plank\textsuperscript{\faCompass}\textsuperscript{\faMountain}\textsuperscript{\faRobot}\\
        \textsuperscript{\faCompass}Department of Computer Science, IT University of Copenhagen, Denmark \\ 
        \textsuperscript{\faMountain}Center for Information and Language Processing (CIS), LMU Munich, Germany \\
     \textsuperscript{\faRobot}Munich Center for Machine Learning (MCML), Munich, Germany \\
        \texttt{elba@itu.dk} \hspace{.5em} \texttt{bplank@cis.lmu.de}}

\begin{document}
\maketitle

\begin{abstract}
Relation Extraction (RE) has attracted increasing attention, but current RE evaluation is limited to in-domain evaluation setups. Little is known on how well a RE system fares in challenging, but realistic out-of-distribution evaluation setups. To address this gap, we propose \projname{}, a new, freely-available cross-domain benchmark for RE, which comprises six distinct text domains and includes multi-label annotations. An additional innovation is that we release \textit{meta-data} collected during annotation, to include explanations and flags of difficult instances. We provide an empirical evaluation with a state-of-the-art model for relation classification. As the meta-data enables us to shed new light on the state-of-the-art model, we provide a comprehensive analysis on the impact of difficult cases and find correlations between model and human annotations. Overall, our empirical investigation highlights the difficulty of cross-domain RE. We release our dataset, to spur more research in this direction.\footnote{ \url{https://github.com/mainlp/CrossRE}}
\end{abstract}

\section{Introduction}

Relation Extraction (RE) is the task of extracting structured knowledge from unstructured text. Although the fact that the task has attracted increasing attention in recent years, there is still a large gap in comprehensive evaluation of such systems which include out-of-domain setups~\cite{bassignana-plank-2022-mean}. 
Despite the drought of research on cross-domain evaluation of RE, its practical importance remains.
Given the wide range of applications for RE to downstream tasks which can vary from question answering, to knowledge-base population, to summarization, and to all kind of other tasks which require extracting structured information from unstructured text, out-of-domain generalization capabilities are extremely beneficial.
It is essential to build RE models that transfer well to new unseen domains, which can be learned from limited data, and work well even on data for which new relations or entity types have to be recognized. 

One direction which is gaining attention is to study RE systems under the assumption that new relation types have to be learned from few examples (\textit{few-shot learning};~\citealp{han-etal-2018-fewrel,gao-etal-2019-fewrel}). One other direction is to study how sensitive a RE system is under the assumption that the input text features change (\textit{domain shift};~\citealp{plank-moschitti-2013-embedding}). There exists a limited amount of studies that focus on the latter aspect, and---to the best of our knowledge---there exists only one paper that proposes to study both, few-shot relation classification under domain shift~\cite{gao-etal-2019-fewrel}. However, this last work considers only two domains---Wikipedia text for training and biomedical literature for testing---and has been criticized for its unrealistic setup~\cite{sabo-etal-2021-revisiting}.

\begin{figure}
\centering
\includegraphics[width=\columnwidth]{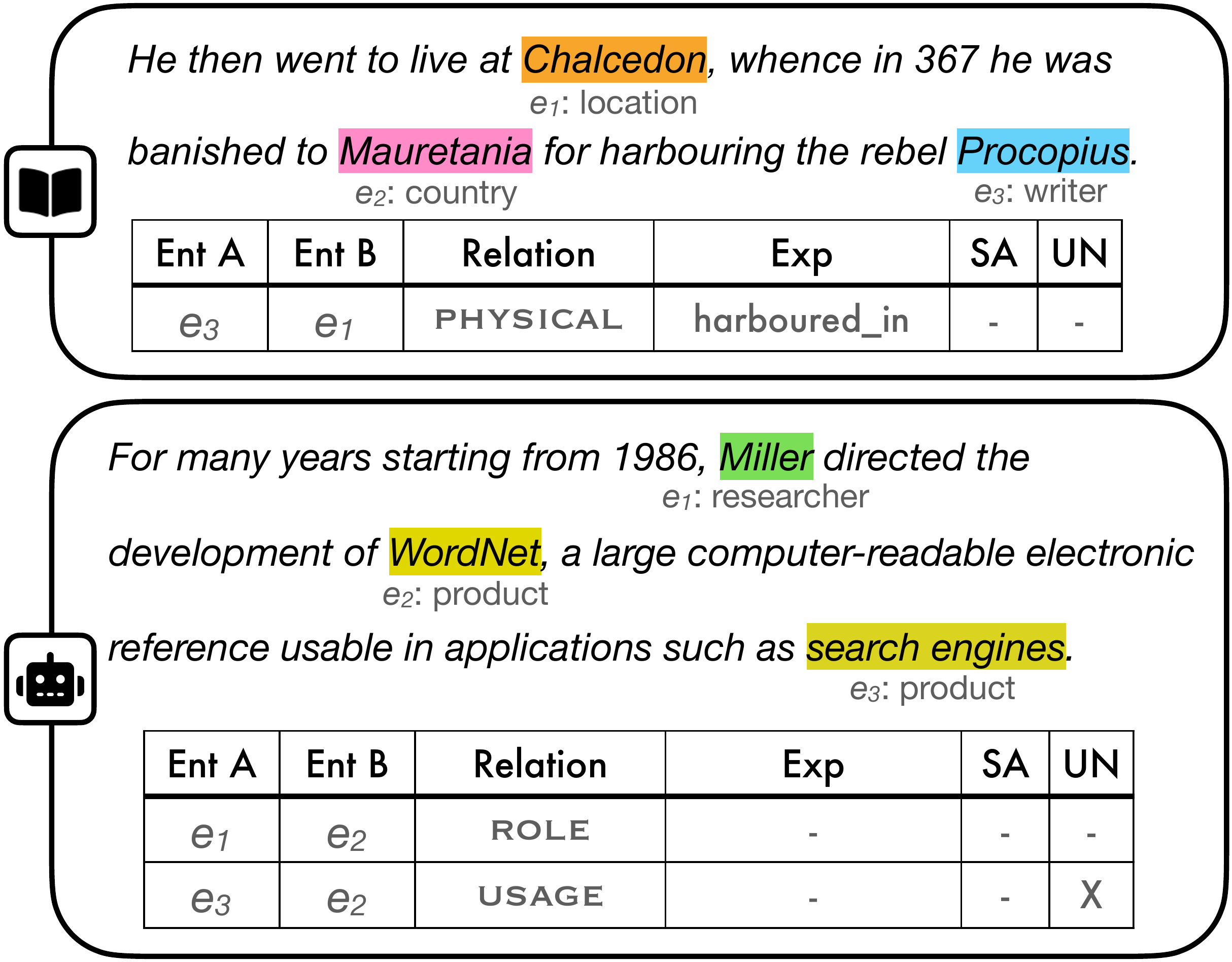}
\caption{\textbf{\projname{} Samples from  Literature and Artificial Intelligence Domains.} At the top, the relation is enriched with the \textsc{Explanation} (Exp) "harboured\_in". At the bottom, instead, the second relation is marked with \textsc{Uncertainty} (UN) by the annotator.}
\label{fig:fig1}
\end{figure}

In this paper, we propose \projname{}, a new challenging cross-domain evaluation benchmark for RE for English (samples in Figure~\ref{fig:fig1}). \projname{} is manually curated with hand-annotated relations covering up to 17 types, and includes multi-label annotations. It contains six diverse text domains, namely:\ news, literature, natural sciences, music, politics and artificial intelligence. One of the challenges of \projname{} is that both entities and relation type distributions vary considerably across domains. \projname{} is heavily inspired by CrossNER~\cite{crossNER}, a recently proposed challenging benchmark for Named Entity Recognition (NER). We extend CrossNER to RE and collect additional meta-data including explanations and flags of difficult instances.
To the best of our knowledge, \projname{} is the most diverse RE datasets available to date, enabling research on domain adaptation and few-shot learning. In this paper we contribute:
\begin{itemize}
    \item A new, comprehensive, manually-curated and freely-available RE dataset covering six diverse text domains and over 5k sentences.
    
    \item We release meta-data collected during annotation, and the annotation guidelines.

    \item An empirical evaluation of a state-of-the-art relation classification model and an experimental analysis of the meta-data provided.
\end{itemize}

\section{Related Work}

Despite the popularity of the RE task~(e.g.\ \citealp{nguyen-grishman-2015-relation,miwa-bansal-2016-end,baldini-soares-etal-2019-matching,wang-lu-2020-two,zhong-chen-2021-frustratingly}), the cross-domain direction has not been widely explored. There are only two datasets which can be considered an initial step towards cross-domain RE.
The ACE dataset~\cite{doddington-etal-2004-automatic} has been analyzed considering its five domains: news (broadcast news, newswire), weblogs, telephone conversations, usenet and broadcast conversations~\cite{plank-moschitti-2013-embedding,nguyen-grishman-2014-employing,nguyen-grishman-2015-event}.
In contrast to ACE, the domains in \projname{} are more distinctive, with specific and more diverse entity types in each of them.

More recently, the FewRel 2.0 dataset~\cite{gao-etal-2019-fewrel}, has been published. It builds upon the original FewRel dataset~\cite{han-etal-2018-fewrel}---collected from Wikipedia---and adds a new test set in the biomedical domain, collected from PubMed.

\section{CrossRE}

\subsection{Motivation}
\label{sec:motivation}

RE aims to extract semantically informative triples from unstructured text. The triples comprehend an ordered pair of text spans which represent named entities or mentions, and the semantic relation which holds between them. The latter is usually taken from a pre-defined set of relation types, which typically changes across datasets, even within the same domain.
The absence of standards in RE leads to models which are designed to extract specific relations from specific datasets. As a consequence, the ability to generalize over out-of-domain distributions and unseen data is usually lacking.
While such specialized models could be useful in applications where particular knowledge is required (e.g.\ the bioNLP field), in most of the cases a more generic level is enough to supply the information required for the downstream task.
In conclusion, RE models that are able to generalize over domain-specific data would be beneficial in terms of both costs of developing and training RE systems designed to work in pre-defined scenarios.
To fill this gap, and in order to encourage the community to explore more the cross-domain RE angle, we publish \projname{}, a new dataset for RE which includes six different domains, with a unified label set of 17 relation types.\footnote{Our data statement~\cite{bender-data-statement} can be found in Appendix~\ref{app:data-statement}.}

\begin{figure*}
\centering
\includegraphics[width=\textwidth]{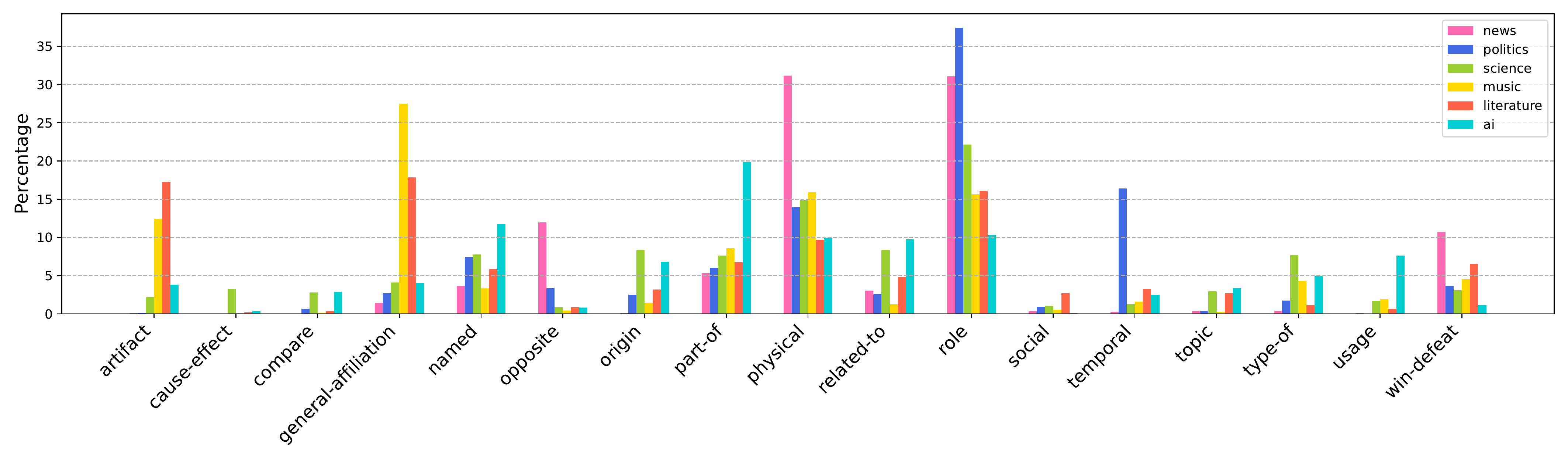}
\caption{\textbf{\projname{} Label Distribution.} Percentage label distribution over the 17 relation types divided by \projname{}'s six domains. Detailed counts and percentages in Appendix~\ref{app:statistics}.}
\label{fig:label-distribution}
\end{figure*}

\subsection{Dataset Overview}
\label{sec:dataset-background}

\projname{} includes the following domains: news (\faNewspaper), politics (\faLandmark), natural science (\faLeaf), music (\faMusic), literature (\faBookOpen) and artificial intelligence (\faRobot; AI). 
Our semantic relations are annotated on top of CrossNER~\cite{crossNER}, a cross-domain dataset for NER which contains domain-specific entity types.\footnote{\url{https://github.com/zliucr/CrossNER/tree/main/ner_data}}
The news domain (collected from Reuters News) corresponds to the data released for the CoNLL-2003 shared task~\cite{tjong-kim-sang-de-meulder-2003-introduction}, while the other five domains have been collected from Wikipedia.
The six domains have been proposed and defined by previous work, and shown to contain diverse vocabularies. We refer to~\citet{crossNER} for details on e.g.\ vocabulary overlap across domains.

During our relation annotation process, we additionally correct some mistakes in named entities previously annotated in CrossNER (entity type, entity boundaries), but only revise existing entity mentions involved in a semantic relation, as well as add new entities involved in semantic relations (see samples in Appendix~\ref{app:data-samples}).

The final dataset statistics are reported in Table~\ref{tab:statistics}.
We keep the train/dev/test data split by~\citet{crossNER} and because of resource constraints, we fix as lower bound the sentence amount of the smallest domain (AI). We pursue their design choice of making training sets relatively small as cross-domain models are expected to do fast adaptation with a small-scale of target domain data samples.
Our annotations are at the sentence-level, and the number of relations indicates the amount of directed entity pairs which are annotated with at least one of the 17 relation labels.

The final dataset contains 17 relation labels for the six domains: \textsc{part-of}, \textsc{physical}, \textsc{usage}, \textsc{role}, \textsc{social}, \textsc{general-affiliation}, \textsc{compare}, \textsc{temporal}, \textsc{artifact}, \textsc{origin}, \textsc{topic}, \textsc{opposite}, \textsc{cause-effect}, \textsc{win-defeat}, \textsc{type-of}, \textsc{named}, and \textsc{related-to}. The latter, very generic, encapsulates all the semantic relations occurring with an extremely low frequency. With this label we make a step forward in respect to~\citet{sabo-etal-2021-revisiting} which merge the `other' and `no-relation' cases into the `None-of-the-above' (NOTA) label. We provide the description of each relation type in Appendix~\ref{app:label-description}, and the full annotation guidelines in our repository. The resulting  label distribution is illustrated in Figure~\ref{fig:label-distribution}, showing that relations vary substantially across domains. We will return to this point in the experimental section and provide further details in the next Section. After that, we describe the process that resulted in the final annotation guidelines and relation types. This includes the details on annotation agreement.

\begin{table}
    \centering
    \resizebox{\columnwidth}{!}{
    \begin{tabular}{c|ccc|c|ccc|c}
    \toprule
    & \multicolumn{4}{c|}{\textsc{sentences}} & \multicolumn{4}{c}{\textsc{relations}} \\
    \midrule
    & train & dev & test & \textbf{tot.} & train & dev & test & \textbf{tot.} \\
    \midrule
    \faNewspaper & 164 & 350 & 400 & 914 & 175 & 300 & 396 & 871 \\
    \faLandmark & 101 & 350 & 400 & 851 & 502 & 1,616 & 1,831 & 3,949 \\
    \faLeaf & 103 & 351 & 400 & 854 & 355 & 1,340 & 1,393 & 3,088 \\
    \faMusic & 100 & 350 & 399 & 849 & 496 & 1,861 & 2,333 & 4,690 \\
    \faBookOpen & 100 & 400 & 416 & 916 & 397 & 1,539 & 1,591 & 3,527 \\
    \faRobot & 100 & 350 & 431 & 881 & 350 & 1,006 & 1,127 & 2,483 \\
    \midrule
    \textbf{tot.} & 668 & 2,151 & 2,446 & \textbf{5,265} & 2,275 & 7,662 & 8,671 & \textbf{18,608} \\
    \bottomrule
    \end{tabular}}
    \caption{\textbf{\projname{} Statistics.} Number of sentences and number of relations annotated for each domain.
    }
    \label{tab:statistics}
\end{table}

As mentioned, our guidelines allow for \emph{multi-label annotations}~\cite{jiang-etal-2016-relation}. 
This means that each entity pair can be assigned to multiple relation types---except for the \textsc{related-to} label which is exclusive and has to be used when none of the other 16 labels fit the data (see example in Appendix~\ref{app:multi-label-sample}).
The combination of labels enables more precise annotations which better represent the meaning expressed in the text (e.g.\ domain-specific scenarios), by keeping the relation label set relatively small and generic, as motivated in Section~\ref{sec:motivation}.
Overall, 6\% of the relations in \projname{} are annotated with multiple labels, specifically: \faNewspaper{} 2\%, \faLandmark{} 15\%, \faLeaf{} 5\%, \faMusic{} 4\%, \faBookOpen{} 2\%, and \faRobot{} 4\%. Note that because of the directionality of the relations, entity pairs containing the same entities, but reverse order, do not count as multi-labeled.

\subsection{Label Distributions}\label{subsec:labeldistr}

\projname{} includes the same label set over its six domains. This implementation choice is motivated by the aim of studying cross-domain RE models which are able to generalize over domain-specific data, and abstract to non-domain-specific relations.
The result is a dataset with divergent label distributions across the different domains. Figure~\ref{fig:label-distribution} shows the label distribution over \projname{}.

From the individual distributions emerges the distinctiveness of each domain.
News includes mainly \textsc{opposite} and \textsc{win-defeat} relations referring to wars, countries being against each other, or sport news about matches between different teams; \textsc{physical}, as many instances include the actual location of the news, and \textsc{role} given that most instances in news are about describing business relationships between organizations or countries.

The politics domain contains \textsc{opposite} and \textsc{win-defeat}, typically political parties and politician being against each other and winning, or losing the elections; the elections, mentioned quite often, usually supply information about the time and so are linked to other entities with the \textsc{temporal} relation. Last, the politics domain presents a high amount of \textsc{role} relations as most of the sentences describe business relations between politicians and political parties or organizations.

Natural science presents a more homogeneous distribution. Distinctively, but similar to AI, which also contain technical text, a higher percentage of relations in respect to the other domains are annotated as \textsc{related-to}, as they would require specialized labels.
Furthermore, similar to AI, the \textsc{origin} label stands out by linking ideas, algorithms, and inventions described in such domains to scientists and researchers. In AI the \textsc{named} relation is also distinctively used, given the wide use in this field of acronyms preceded by their extension.

Last, music and literature have a particular high number of \textsc{artifact} labels describing songs, albums and books made and written by bands, musicians and writers, and \textsc{general-affiliation} relations linking songs, albums, musicians, books and writers to specific music and literary genres.

\subsection{Annotation Guidelines Definition Process}
\label{sec:guideline-definition}

We bootstrap the dataset starting with a traditional top-down process, using an initial set of existing labels~\cite{doddington-etal-2004-automatic,hendrickx-etal-2010-semeval,gabor-etal-2018-semeval,luan-etal-2018-multi}, but continue by following a bottom-up approach (\emph{data-driven annotation}), with the goal to annotate all the semantic relations present in the data, while balancing a trade-off between specificity (to domain-specific labels) and generalizability~\cite{pustejovsky2012natural}. The whole process (annotation guideline definition and data  annotation) lasted around seven months, and is depicted next.

The guidelines have been defined via an iterative process including a total of seven annotation rounds (two preliminary and five official rounds).
The two preliminary rounds have been completed by in-house NLP experts, with one round in the entire lab.
The latter has been particularly crucial for collecting different points of view about the relations present in the dataset.
After those, a hired expert with a linguists degree (who is the official annotator of the dataset) entered the process and the five official rounds began. These last rounds have been performed by the linguist together with one NLP expert, in consultation with a third NLP expert during the plenary discussion rounds. 

The annotators in the official rounds were allowed to use the labels from the defined set, and were asked to explain their choice with a more fine-grained type (written in free text, typically as a predicate like `won\_award'). In addition, they were initially allowed to define new relation labels if a case was not fitting in any of the proposed ones.
Each annotation round was carried out individually by each annotator and was followed by a plenary discussion. During the latter the given guidelines were reviewed and modified for the next annotation round. The process continued until the current high annotation agreement was achieved (see Section~\ref{sec:annotation-agreement}), after which the professional annotator continued to annotate the rest. This took close to 5 months of near full-time (0.8 fte) employment.

\begin{table*}[]
\resizebox{\textwidth}{!}{
\centering
\begin{tabular}{l}
\toprule
\textbf{\textsc{Explanation} (\textsc{Exp})} \\
\hline
\begin{dependency}
\begin{deptext}
\emph{On 12 April 2019 a new Eurosceptic party, the} \& \emph{Brexit Party} \& \emph{was officially launched by former} \& \emph{UK Independence Party} \& \emph{Leader} \& \emph{Nigel Farage}\&. \\ \\
\& $e_1$: political party \& \& $e_2$: political party \& \& $e_3$: politician \& \\
\end{deptext}
\wordgroup[group style={fill=orange!40, draw=brown}]{1}{2}{2}{entity1}
\wordgroup[group style={fill=orange!40, draw=brown}]{1}{4}{4}{entity2}
\wordgroup[group style={fill=blue!85!yellow!30!white, draw=blue!60!black}]{1}{6}{6}{entity3}
\end{dependency}\\
($e_1$, $e_3$, \textsc{origin}, \textsc{Exp}: founded\_by) ($e_3$, $e_1$, \textsc{role}, \textsc{Exp}: founder\_of) ($e_3$, $e_2$, \textsc{role}, \textsc{Exp}: former\_leader\_of) \\
\midrule
\textbf{\textsc{Syntax Ambiguity} (\textsc{SA})} \\
\hline
\begin{dependency}
\begin{deptext}
\emph{Variants of the} \& \emph{back-propagation algorithm} \& \emph{as well as} \& \emph{unsupervised methods} \&
\emph{by} \& \emph{Geoff Hinton} \& \emph{and colleagues at the} \& \emph{University of Toronto} \& \emph{can be used [...]}
\\ \\
\& $e_1$: algorithm \& \& $e_2$: misc \& \& $e_3$: researcher \& \& $e_4$: university \& \\
\end{deptext}
\wordgroup[group style={fill=orange!40, draw=brown}]{1}{2}{2}{entity1}
\wordgroup[group style={fill=yellow!40, draw=brown!50!yellow}]{1}{4}{4}{entity2}
\wordgroup[group style={fill=blue!85!yellow!30!white, draw=blue!60!black}]{1}{6}{6}{entity3}
\wordgroup[group style={fill=green!85!blue!30!white, draw=green!60!black}]{1}{8}{8}{entity4}
\end{dependency} \\
($e_1$, $e_3$, \textsc{origin}, \textsc{SA}: True) ($e_2$, $e_3$, \textsc{origin}) ($e_3$, $e_4$, \textsc{role}) ($e_3$, $e_4$, \textsc{physical}) \\
\midrule
\textbf{\textsc{Uncertainty} of the annotator (\textsc{Un})} \\
\hline
\begin{dependency}
\begin{deptext}
\emph{DNA methyltransferase} \& \emph{is recruited to the site and adds} \& \emph{methyl groups} \& \emph{to the} \& \emph{cytosine} \& \emph{of the} \& \emph{CpG dinucleotides}\&.
\\ \\
$e_1$: enzyme \& \& $e_2$: misc \& \& $e_3$: chemical compound \& \& $e_4$: misc \& \\
\end{deptext}
\wordgroup[group style={fill=orange!40, draw=brown}]{1}{1}{1}{entity1}
\wordgroup[group style={fill=yellow!40, draw=brown!50!yellow}]{1}{3}{3}{entity2}
\wordgroup[group style={fill=blue!85!yellow!30!white, draw=blue!60!black}]{1}{5}{5}{entity3}
\wordgroup[group style={fill=yellow!40, draw=brown!50!yellow}]{1}{7}{7}{entity4}
\end{dependency} \\
($e_1$, $e_2$, \textsc{related-to}, \textsc{Un}: True) ($e_2$, $e_3$, \textsc{part-of}, \textsc{Un}: True) ($e_3$, $e_4$, \textsc{part-of}, \textsc{Un}: True) \\
\bottomrule
\end{tabular}
}
\caption{\label{tab:metadata-samples}
\textbf{Samples of Meta-data Annotations.} Annotation samples from \projname{} which have been enriched with meta-data: \textsc{Explanation} of the relation type assigned, \textsc{Syntax Ambiguity} which poses a challenge for the annotator, and \textsc{Uncertainty} of the annotator.
}
\end{table*}

\subsection{Annotation Agreement}
\label{sec:annotation-agreement}

With the aim of a more fine-grained analysis of the annotation agreement, we split RE into its two task components: Relation Identification (RI) and Relation Classification (RC).
The first is the identification task which given a sentence and two marked entities determines if there exist one of the 17 semantic relation between them. The second, more fine-grained, takes the positive sample from RI and, given the label set, classifies the instances into the specific relation types.
Such division supported the guideline definition process in order to understand whether the label descriptions were not specific enough, or whether there was unclarity in detecting the presence of a relation at all.

\begin{figure}
\centering
\includegraphics[width=\columnwidth]{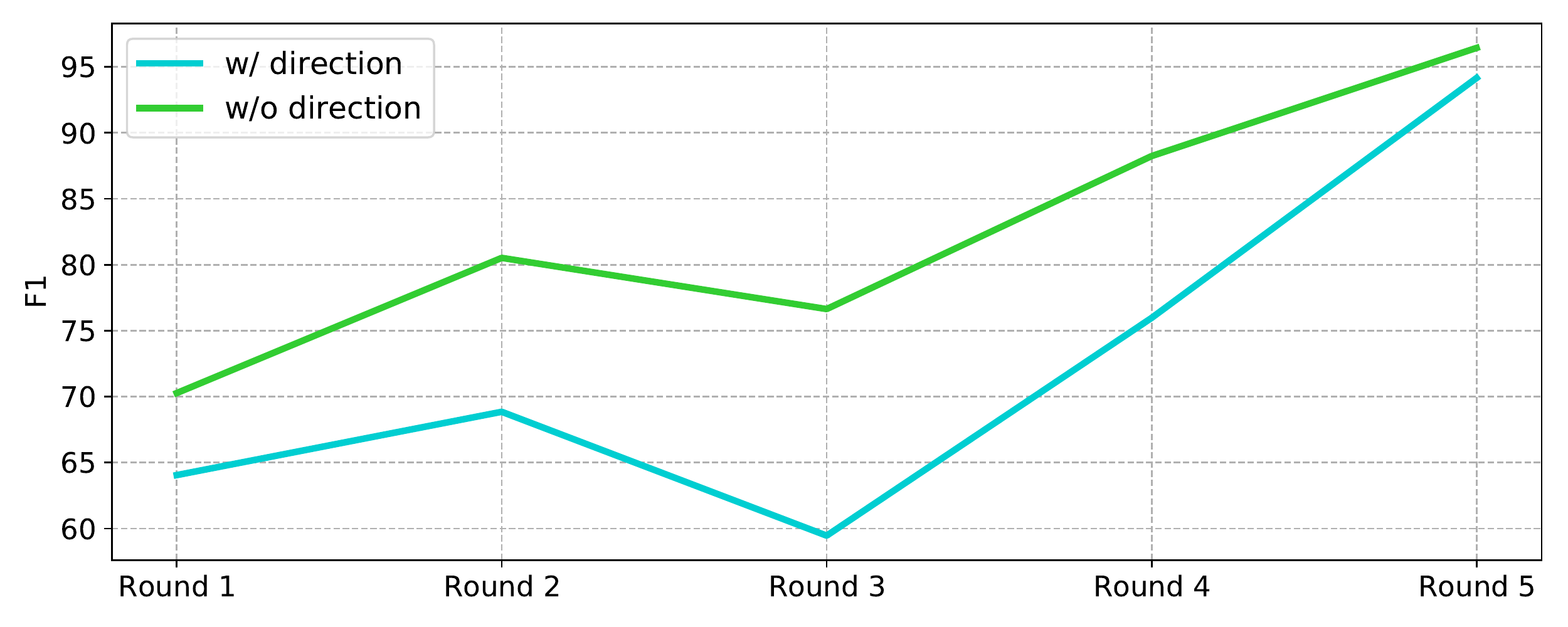}
\caption{\textbf{RI Annotation Agreement.} F1 score of the identified relations during the official annotation rounds.}
\label{fig:ri-agreement}
\end{figure}

\begin{figure}
\centering
\includegraphics[width=\columnwidth]{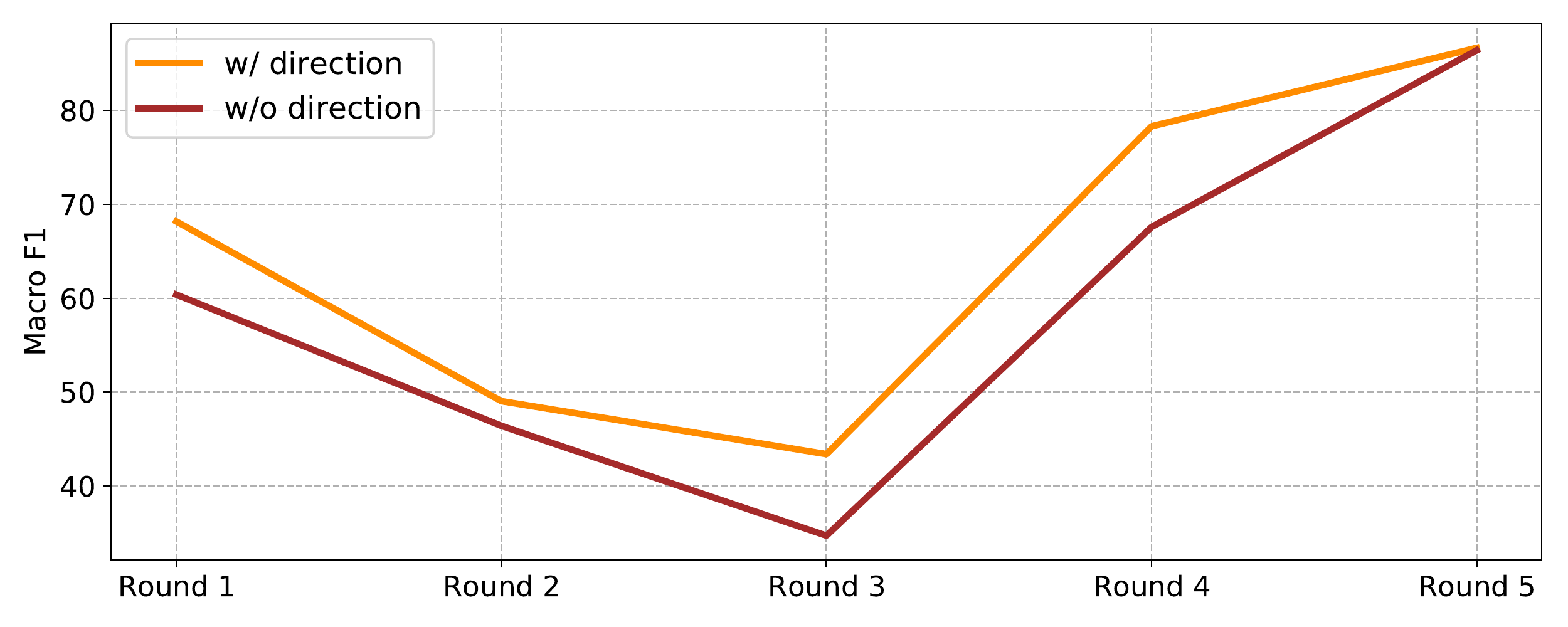}
\caption{\textbf{RC Annotation Agreement.} Macro-F1 score of the assigned labels over the entity pairs identified by both annotators during the official annotation rounds.}
\label{fig:rc-agreement}
\end{figure}

As described in Section~\ref{sec:guideline-definition}, the guideline definition has been an iterative process with five annotation rounds and Figures~\ref{fig:ri-agreement} and \ref{fig:rc-agreement} report the annotation agreement between the linguist and the NLP expert. As the entity order is part of the annotation guidelines, we furthermore tease apart the directionality component for a deeper analysis of the annotation agreement.

In Figure~\ref{fig:ri-agreement} we see that when considering the direction---$(e_1,e_2) \neq (e_2,e_1)$---the RI agreement is lower as we are considering one additional constraint in respect to the looser setup where $(e_1,e_2) = (e_2,e_1)$. In Figure~\ref{fig:rc-agreement} RC presents, instead, an inverse trend which is motivated by the fact that if the annotators agree on the direction, they will more likely assign the same relation label.

Several interesting observation emerge during the process. First, the drop in round 2 for RC indicates that it was at first easier to identify a relation between two entities  (as RI agreement increases) than determining the exact label (RC agreement decreases). Therefore, between round 2 and 3 the discussion was centered around specifying the relation type descriptions and their respective directionality in more detail. The effect of this is visible in the next rounds, which resulted at first in an annotation agreement drop for RI (and consequently slight drop in RC agreement), but starting from round 3 onwards we observe a steady increase: This is also the point that marked the final version of the annotation guidelines, which remained stable and the annotators were trained to use them over rounds 3, 4, 5.
The converging agreements (w/ and w/o direction) of round 5 for both RI and RC indicate that the annotators achieved high data quality, annotating relations correctly.

The last annotation round (Round 5) included 72 sentences (12 from each of the six domains) for a total of 2,284 tokens resulting in high agreement. In particular,  RI agreement considering the direction of the entity pairs is 94.16 F1 and without considering it 96.44 F1. The RC agreement considering the direction is 86.65 Macro-F1, and without considering it 86.39 Macro-F1. Furthermore, as we check and correct the entity spans from the previous NER datasets (see Section~\ref{sec:dataset-background}), we additionally compute the entity annotation agreement. Regarding entities, the Span-F1 with respect to the original data source is 90.79 and 91.81 respectively for the official annotator and the NLP expert, while the Span-F1 between them increases to 94.43, indicating that there is high consistency in correcting the entities.
In light of the increasing interest to question the  strong assumption of one unique gold label~\citep{plank-etal-2014-linguistically, basile-etal-2021-need}, we also release the doubly-annotated data from the last round in our repository to spurge research on learning with human label variation.

\subsection{Meta-data Annotation}
\label{sec:innovations}

By embracing the subjectivity of manually-curated datasets, we collect \emph{meta-data} (see data samples in Table~\ref{tab:metadata-samples}). We hope this facilitates future analyses of the dataset, including new annotation iterations, and interpretability of the predictions.

We include an \textsc{Explanation} field for adding notes or specifications regarding the label assigned. In the first example in Table~\ref{tab:metadata-samples}, the first relation (\textsc{origin}) is motivated by $e_1$ having been founded by $e_3$. Similarly the second relation, which includes the same entities, but with inverse order given the directionality of the \textsc{role} label---note that this is not counted as multi-labeled as also the order has to match. In the last triple \textsc{role} assumes a different meaning and it is specified in the \textsc{Exp} field by `former\_leader\_of'.
Furthermore, we include two check-boxes. One is for identifying the presence of \textsc{Syntax Ambiguity}, which poses a challenge for the annotator.
In the second example in Table~\ref{tab:metadata-samples}, while we can confidently state that $e_2$ has been originated by $e_3$, the scenario for $e_1$ is ambiguous, and therefore the first triple is marked with `\textsc{SA}: True'.
The other check-box, named \textsc{Uncertainty}, allows the indication of low confidence by the annotator on the relation identified or on the label assigned. For instance, the third example in Table~\ref{tab:metadata-samples} (from the science domain) contains technical text which may require deeper knowledge of an expert in the field, and so our annotator (a linguist) flagged the relations in it as \textsc{Uncertainty}. 
The meta-data described have been extremely useful for the guideline definition process.

Table~\ref{tab:metadata-statistics} reports the statistics of the meta-data annotations.
The domains where our annotator is less confident are natural science and AI, and these are also the ones which contain more technical text specific to the two respective fields.

\begin{table}[]
    \centering
    \resizebox{\columnwidth}{!}{
    \begin{tabular}{r|cccccc|c}
        \toprule
         & \faNewspaper & \faLandmark & \faLeaf & \faMusic & \faBookOpen & \faRobot & \textbf{tot.} \\
        \midrule
        \textsc{Exp} & 138 & 479 & 421 & 777 & 1,036 & 448 & 3,299 \\
        \textsc{SA} & 0 & 32 & 20 & 169 & 31 & 25 & 277 \\
        \textsc{Un} & 6 & 17 & 126 & 23 & 37 & 238 & 447 \\
        \bottomrule
    \end{tabular}}
    \caption{\textbf{Meta-data Statistics.} Amount of annotations which have been marked with the following metadata: \textsc{Explanation} (\textsc{Exp}), \textsc{Syntax Ambiguity} (\textsc{SA}), and \textsc{Uncertainty} of the annotator (\textsc{Un}). The counts refer to the sum over train, dev, and test.}
    \label{tab:metadata-statistics}
\end{table}

\section{Baseline Experiments}

We provide the evaluation of a state-of-the-art model on the proposed dataset.
To establish baselines, we train models over each of the proposed domains.
Two major challenges affecting the dataset are the multi-label annotation setup and the highly sparse label distribution distinctive of each domain.

\subsection{Experimental Setup}
Within this first empirical evaluation of \projname{}, and given the challenges highlighted above, we follow previous work~\cite{han-etal-2018-fewrel,baldini-soares-etal-2019-matching,gao-etal-2019-fewrel} and focus on Relation Classification (RC) only, leaving the complete RE task for future work.
The goal of RC is to assign the correct relation types to the ordered entity pairs which have been identified as being semantically connected.

\subsection{Model}
\label{sec:model}

Our RC model follows the current state-of-the-art by~\citet{baldini-soares-etal-2019-matching}.
Given a sentence $s$ and an ordered pair of entity mentions $(e_1,e_2)$, we augment $s$ with four entity markers $e_1^{start}$, $e_1^{end}$, $e_2^{start}$, $e_2^{end}$ which delimit the start and end of the entity spans. Following~\citet{zhong-chen-2021-frustratingly} we enrich the entity markers with information about the entity types.
For example, given the following sentence $s$ and entity mention pair $(e_1,e_2)$:
\begin{dependency}
\begin{deptext}
\emph{Cunningham} \& \emph{played his entire 11-year}\\
$e_1$: person \&  \\
\emph{career with the} \& \emph{Philadelphia Eagles}\& \\
\& $e_2$: organization \\
\end{deptext}
\wordgroup[group style={fill=orange!40, draw=brown}]{1}{1}{1}{entity1}
\wordgroup[group style={fill=green!85!blue!30!white, draw=green!60!black}]{3}{2}{2}{entity1}
\end{dependency}

$s$ is augmented as:
\begin{quote}
<E1:person> \emph{Cunningham} </E1:person> \emph{played his entire 11-year career with the} <E2:organization> \emph{Philadelphia Eagles} </E2:organization> 
\end{quote}
The above version of $s$ is then fed into a pre-trained encoder (BERT;~\citealp{devlin-etal-2019-bert}) and we denote the output representation by $\hat{s}$.
The output representations of the two start markers are concatenated in $[\hat{s}_{e_1^{start}}, \hat{s}_{e_2^{start}}]$
and used for the relation type classification via a feed-forward neural network.
Given a set of $n$ relation labels, the latter consists of a linear layer with output size $n$, followed by a softmax activation function.
Considering the amount of multi-labeled instances being only around the 6\% over the whole dataset, ignoring them by using a single-head model which can be trained more easily resulted in the best choice.\footnote{We previously tested a multi-head model for enabling multi-label predictions, but the per-label data is not enough to effectively train each of the head classifier.}
We run our experiments over five random seeds.
See Appendix~\ref{app:reproducibility} for hyperparameters settings.

\begin{table}[]
    \centering
    \resizebox{\columnwidth}{!}{
    \begin{tabular}{c|cccccc|c}
        \toprule
         & \faNewspaper & \faLandmark & \faLeaf & \faMusic & \faBookOpen & \faRobot & \textbf{avg.} \\
        \midrule
        \textsc{Micro-F1} &
        46.36 & 58.26 & 40.10 & 75.96 & 67.70 & 45.40 & 55.63 \\
        \midrule
        \textsc{Macro-F1} &
        16.52 & 20.33 & 25.29 & 39.19 & 37.74 & 30.66 & 28.29 \\
        \midrule
        \textsc{Weigh.-F1} &
        37.59 & 53.53 & 35.84 & 73.16 & 63.08 & 41.52 & 50.79 \\
        \midrule
    \end{tabular}}
    \caption{\textbf{\projname{} Baselines.} Results achieved by our baseline model on the RC task. Reported are the averages over five random seeds (see Table~\ref{tab:reproducibility}).}
    \label{tab:baseline-scores}
\end{table}

\subsection{Results}

\paragraph{Evaluation}
For a better evaluation of the baseline, given the highly imbalanced label distributions of the six domains, we follow~\citealp{harbecke-etal-2022-micro} and compute the micro-averaged F1, as well as the macro-averaged F1 and the weighted F1. The macro-average does not consider the classes with a support set of 0 in the test set.\footnote{Evaluation code in our repository.} The per-class data scarcity of most of the labels over the different domains (see Table~\ref{tab:label-distribution}) means the Macro-F1 is lower with respect to the other two metrics. However, it provides a more realistic scenario of the per-class performance of the model, and of the difficulty that the sparsity of the relation types adds in an already challenging classification task with 17 labels.

\paragraph{General Scores}
Table~\ref{tab:baseline-scores} reports the scores achieved by our RC model.
The news domain is the only one based on CoNLL-2003 as opposed to the other five domains (CrossNER).
The instances are mostly news headlines or very short news reports and so, even if the amount of annotated sentences is comparable with the other domains, the semantic relations present in these data are considerably fewer (see Table~\ref{tab:statistics}). This, in addition to the most imbalanced label distribution---predominantly \textsc{role}, \textsc{physical}, \textsc{opposite}, \textsc{win-defeat} and \textsc{part-of} (see Figure~\ref{fig:label-distribution})---leads news to be one the most challenging domain in term of Macro-F1. In contrast, the music domain, with the highest amount of annotated relations, achieves the highest scores in respect to the other domains.

\begin{table}[]
    \centering
    \resizebox{\columnwidth}{!}{
    \begin{tabular}{r|cccccc}
    \toprule
         & \faNewspaper & \faLandmark & \faLeaf & \faMusic & \faBookOpen & \faRobot \\
    \midrule
    \textsc{artifact} & - & 0.0 & 17.93 & 85.74 & 86.13 & 52.55 \\
    \textsc{cause-effect} & - & - & 0.0 & 0.0 & 0.0 & 0.0 \\
    \textsc{compare} & - & 0.0 & 45.39 & 0.0 & 0.0 & 0.0 \\
    \textsc{gen.-aff.} & 0.0 & 24.49 & 29.19 & 87.04 & 84.46 & 3.07 \\
    \textsc{named} & 34.67 & 54.56 & 53.53 & 10.3 & 48.66 & 65.11 \\
    \textsc{opposite} & 9.38 & 4.41 & 0.0 & 0.0 & 2.67 & 0.0 \\
    \textsc{origin} & - & 0.0 & 26.7 & 32.79 & 0.0 & 42.51 \\
    \textsc{part-of} & 0.0 & 2.2 & 19.71 & 38.33 & 11.06 & 49.11 \\
    \textsc{physical} & 45.8 & 71.12 & 73.06 & 91.23 & 76.43 & 76.79 \\
    \textsc{related-to} & 0.0 & 0.0 & 41.51 & 11.81 & 8.98 & 27.44 \\
    \textsc{role} & 58.84 & 59.67 & 40.15 & 65.57 & 63.38 & 61.56 \\
    \textsc{social} & - & 0.0 & 0.0 & 0.0 & 50.34 & 0.0 \\
    \textsc{temporal} & 0.0 &  85.72 & 0.0 & 32.68 & 63.45 & 51.66 \\
    \textsc{topic} & - & 0.0 & 1.14 & 0.0 & 9.48 & 30.73 \\
    \textsc{type-of} & - & 0.0 & 6.57 & 79.29 & 59.73 & 18.68 \\
    \textsc{usage} & - & - & 0.0 & 56.15 & 0.0 & 12.2 \\
    \textsc{win-defeat} & 0.0 & 2.77 & 75.06 & 75.31 & 76.75 & 29.78 \\
    \bottomrule
    \end{tabular}}
    \caption{\textbf{Per-class Results.} Detailed F1 scores for each relation type. Reported are the averages over five random seeds (see Table~\ref{tab:reproducibility}). `-' indicates the class is not present in the test set.}
    \label{tab:per-label-scores}
\end{table}

\paragraph{Per-label Performance}
In Table~\ref{tab:per-label-scores} we report the per-label F1 scores for a more detailed analysis.
Several labels have just few samples in the training sets and so are very difficult to learn, leading to an F1 of 0.0. These cases push down the Macro-F1 scores in Table~\ref{tab:baseline-scores}.
Overall, the amount of instances per-label---see Figure~\ref{fig:label-distribution} for percentages and Table~\ref{tab:label-distribution} for counts---are good indicators for the individual scores in Table~\ref{tab:per-label-scores}.
For example \textsc{general-affiliation} achieves high F1 both in the music and in the literature domains (87.04 and 84.46 respectively). This is similar in \textsc{temporal} in the politics domain (85.72). 
However, we notice that some labels are more challenging than others: While the \textsc{role} label contains more instances than the \textsc{temporal} one in the politics domain, it only achieves a score of 59.67. Given the imbalanced train/dev/test split over the six domains, and in order to give a more realistic idea of the distributions, we report as an example the label distribution over the train/dev/test split of the politics domain in Appendix~\ref{app:politics-distribuition}.
We additionally notice that the same label can have different levels of challenge depending on the domain. For example, \textsc{named} corresponds to similar percentages in the domains of news and music (3.62\% and 3.34\% respectively), but given the disparate total amount of in-domain relations these correspond to very different amounts: 32 in news and 164 in music. However, the \textsc{named} label achieves an F1 score of 34.67 in the news domain, and only 10.3 in the music domain.

\section{Meta-data Analysis}

In this section, we use the meta-data collected during the annotation of the dataset for further analysis.
We consider \textsc{Syntax Ambiguity} (\textsc{SA}) and \textsc{Uncertainty} of the annotator (\textsc{Un}) and examine the performance of our baseline model on such instances.
Table~\ref{tab:test-statistics} reports the meta-data statistics on the six test sets. 
Given the almost absence of samples in the news domain, we leave it out from this analysis.
Table~\ref{tab:meta-data-analysis} shows the results of our model when evaluated on samples only with \textsc{SA} and \textsc{Un}, both, or none, compared to \textsc{all}. For this ablation study we do not report the Macro-F1 because changing the evaluation set would mislead the analysis (as mentioned, the Macro-F1 only considers classes present in the evaluation set).

\begin{table}[]
    \centering
    \resizebox{0.8\columnwidth}{!}{
    \begin{tabular}{r|cccccc}
        \toprule
         & \faNewspaper & \faLandmark & \faLeaf & \faMusic & \faBookOpen & \faRobot \\
        \midrule
        \textsc{SA} &
        0 & 15 & 8 & 150 & 6 & 20 \\
        \textsc{Un} &
        1 & 9 & 62 & 19 & 8 & 68 \\
        \textsc{SA} or \textsc{Un} &
        1 & 23 & 69 & 167 & 12 & 88 \\
        \midrule
    \end{tabular}}
    \caption{\textbf{Test Set Statistics.} Amount of annotations which have been marked with \textsc{Syntax Ambiguity} (\textsc{SA}) and with \textsc{Uncertainty} (\textsc{Un}) in the test sets.}
    \label{tab:test-statistics}
\end{table}

With the low amount of instances in politics an literature, results are less pronounced and differences with the overall scores are absent in most cases. Therefore, we focus here on the remaining three domains---natural science \faLeaf{}, music \faMusic{}, AI \faRobot{}. We observe slightly but consistently higher scores when taking out the cases marked with \textsc{Un}, showing that they are challenging not only for the human but also the system. Those are the cases identified as most challenging, specially considering the annotator's background (i.e.\ natural science and AI, mostly on \textsc{cause-effect}, \textsc{part-of}, \textsc{usage}).
The results in respect to the \textsc{SA} annotations are mixed: There is not a unified trend over domains or metrics. We attribute this to the fact that our model does not explicitly build upon syntactic features (e.g.\ syntactic trees;~\citealp{plank-moschitti-2013-embedding}).
Finally, the scores from the data which consider the combination of \textsc{SA} and \textsc{Un} increase over the baseline in the science domain, where taking out both \textsc{SA} and \textsc{Un} individually increase over the \textsc{all} setup. 
In the music domain, where \textsc{SA} are frequent, excluding them result in a little drop of Micro-F1 (75.96$\rightarrow$74.67). In fact, the model is good on \textsc{SA} in the music domain: The majority of cases are on the \textsc{general-affiliation} label, which achieves high per-label F1 (87.04). We attribute it to the fact that in this domain there are many lists of entities and relative attributes, which structurally can be ambiguous, but often involve a similar relation structure. AI presents a similar trend as music, but the scores from the combination of \textsc{SA} and \textsc{Un} increase a bit in both metrics.

\begin{table}[t!]
    \centering
    \resizebox{\columnwidth}{!}{
    \begin{tabular}{c|rr|ccccc}
        \toprule
        & & & \faLandmark & \faLeaf & \faMusic & \faBookOpen & \faRobot \\
        \midrule
        \multirow{7}{*}{\rotatebox{90}{\textsc{Micro F1}}} & 
        \multicolumn{2}{c|}{\textsc{all}} & 58.26 & 40.10 & 75.96 & 67.70 & 45.40 \\
        \cmidrule(r){2-8}
        & \multirow{2}{*}{\textsc{SA}} &
        w/ & 54.67 & 12.50 & 95.25 & 76.67 & 87.00 \\
        & & w/o & 58.27 & 40.26 & 74.67 & 67.66 & 44.68 \\
        \cmidrule(r){2-8}
        & \multirow{2}{*}{\textsc{Un}} &
        w/ & 66.67 & 20.97 & 27.00 & 67.50 & 27.34 \\
        & & w/o & 58.21 & 40.97 & 76.38 & 67.70 & 46.58 \\
        \cmidrule(r){2-8}
        & \textsc{SA or} &
        w/ & 57.39 & 20.29 & 87.04 & 70.00 & 40.67 \\
        & \textsc{Un}& w/o & 58.26 & 41.11 & 75.12 & 67.68 & 45.82 \\
        \midrule
        \multirow{7}{*}{\rotatebox{90}{\textsc{Weighted F1}}} & 
        \multicolumn{2}{c|}{\textsc{all}} & 53.53 & 35.84 & 73.16 & 63.08 & 41.52 \\
        \cmidrule(r){2-8}
        & \multirow{2}{*}{\textsc{SA}} &
        w/ & 57.16 & 13.33 & 94.91 & 74.57 & 87.00 \\
        & & w/o & 53.62 & 36.00 & 71.66 & 63.06 & 40.97 \\
        \cmidrule(r){2-8}
        & \multirow{2}{*}{\textsc{Un}} &
        w/ & 61.78 & 12.91 & 30.46 & 64.46 & 19.13 \\
        & & w/o & 53.50 & 36.68 & 73.59 & 63.11 & 42.95 \\
        \cmidrule(r){2-8}
        & \textsc{SA or} &
        w/ & 55.62 & 13.18 & 86.79 & 64.24 & 32.43 \\
        & \textsc{Un}& w/o & 53.62 & 36.83 & 72.13 & 63.10 & 42.38 \\
        \bottomrule
    \end{tabular}}
    \caption{\textbf{Meta-data Analysis.} F1 scores on the instances which have been marked with \textsc{Syntax Ambiguity} (\textsc{SA}), \textsc{Uncertainty} (\textsc{Un}), or at least one of the two. We report also the baselines of Table~\ref{tab:baseline-scores} (\textsc{all}).}
    \label{tab:meta-data-analysis}
\end{table}

In conclusion, we do gain informative insights from the collected meta-data---especially when the annotator is unsure about the annotated relation, and also to understand whether syntactic ambiguity detected by the annotator impacts system accuracy.

\section{Conclusion}

We present \projname{}, a new challenging manually-curated corpus for RE. It is the first dataset for RE covering six diverse text domains (news, politics, natural science, music, literature, AI) with annotations spanning 17 relation types. 
Some annotations are enriched with meta-data information (explanation for the choice of the assigned label, identification of syntax ambiguity, and uncertainty of the annotator). Throughout the annotation process and in the empirical validation, this meta-data proves to be useful and insightful. As it aids the analysis of the provided baseline, we invite the research community to both collect and release such additional information.

We perform an empirical evaluation of \projname{} by applying state-of-the-art RC methods~\cite{baldini-soares-etal-2019-matching,zhong-chen-2021-frustratingly}, and show the challenges of its highly imbalanced label distributions over the domains.

The cross-domain dimension is currently under-explored in the RE field. With this dataset we invite future work on cross-domain RE evaluation, the exploration of domain-adaptive techniques (e.g.\ DAPT; \citealp{gururangan-etal-2020-dont}) and other adaptation methods to improve the baseline set out in this work for the different data domains.

\section*{Limitations}

Because of resource constraints (time and costs, see next Section), the proposed dataset is limited to one annotator. However, as our annotation process details show, we expect the quality to be high, nevertheless, preferably if resources were available, gaining larger subsets with multiple annotations would be a promising next step.
Crucially, we involved the annotator in the guideline definition process, which was very fruitful and inspired us to collect syntax ambiguity information as well.

We identify as a second limitation the fact that five out of six of our domains belong to the same data source (Wikipedia). However, the advantage is that Wikipedia data can be redistributed freely. We acknowledge the already challenging setup of our dataset, but invite future work on the inclusion of different data sources whenever possible.

\section*{Ethics Statement}

The data included in our newly proposed dataset correspond to a sub-set of the data collected and freely published by~\citet{crossNER} within the CrossNER project.

Our dataset is annotated by a hired expert with a linguists degree employed on 0.8fte for this project following national  salary rates. The total costs for data annotation amount to roughly 19,000 USD, amounting to $\approx$ 1\$ per annotated relation.

\section*{Acknowledgements}
First, we would like to thank our annotator for the great job and substantial help given to this project.
We thank Filip Ginter and Sampo Pyysalo for insightful discussion at the last stage of this work. 
Furthermore, we thank the NLPnorth group for feedback on an
earlier version of this paper---in particular Rob van der Goot, Mike Zhang, and  Max Müller-Eberstein. 
Last, we also thank the ITU's High-performance Computing cluster for computing resources. 
Elisa Bassignana and Barbara Plank are supported by the Independent Research Fund Denmark (Danmarks Frie Forskningsfond; DFF) Sapere Aude grant 9063-00077B. Barbara Plank is supported by the ERC Consolidator Grant DIALECT  101043235.

\bibliography{anthology,custom}
\bibliographystyle{acl_natbib}

\appendix

\section{Data Statement \projname{}}
\label{app:data-statement}

Following~\citep{bender-data-statement} we outline below the data statement fo \projname{}.

\begin{itemize}
    \item[A.] \textsc{Curation Rationale:}
    Collection of Reuters News and Wikipedia pages annotated with the aim of studying Relation Extraction.
    \item[B.] \textsc{Language Variety:}
    The language is English. For additional details we refer to~\citealp{tjong-kim-sang-de-meulder-2003-introduction} and to~\citealp{crossNER} who did the data collection.
    \item[C.] \textsc{Speaker Demographic:}
    Unknown.
    \item[D.] \textsc{Annotator Demographic:}
    One sprofessional annotator with a background in Linguistics and one NLP expert with a background in Computer Science. Age range: 25–30; Gender: both female; Race/ethnicity: white European; Native language: Danish, Italian; Socioeconomic status: higher-educated.
    \item[E.] \textsc{Speech Situation:}
    We refer to~\citealp{tjong-kim-sang-de-meulder-2003-introduction} and to~\citealp{crossNER}.
    \item[F.] \textsc{Text Characteristics:}
    The texts are news from Reuters News, and Wikipedia pages about politics, natural science, literature, artificial intelligence.
    \item[G.] \textsc{Recording Quality:}
    N/A
    \item[H.] \textsc{Other:}
    N/A
    \item[I.] \textsc{Provenance Appendix:}
    The data statements of the previous datasets~\cite{tjong-kim-sang-de-meulder-2003-introduction,crossNER} 
    are not available.
\end{itemize}

\section{Relation Label Description}
\label{app:label-description}
Below we report the description of each relation type we use to annotate \projname{}. We refer to our repository for the complete annotation guidelines, including directionality of the relations, samples. and instruction on what to annotate.

\begin{itemize}
    \item \textsc{part-of} Something that is part of something else (e.g.\ song\_part\_of\_album, task\_ part\_of\_field).
    \item \textsc{physical} Answer the question \textit{Where?} (e.g.\ location, near, destination, located\_in, based\_in, residence, released\_in, come\_from).
    \item \textsc{usage} Something which make use of something else in order to accomplish its scope, includes an agent using an instrument.
    \item \textsc{role} Two entities which are linked by a \textit{business related} role (e.g.\ management, founder, affiliate\_partner, member\_of, citizen\_of, participant, nominee\_of).
    \item \textsc{social} Two entities linked by a \textit{non-business related} role (e.g.\ parent, sibling, spouse, friend, acquaintance).
    \item \textsc{general-affiliation} Religion, ethnicity, genre (e.g.\ book\_genre, music\_genre).
    \item \textsc{compare} Something that is compared with something else.
    \item \textsc{temporal} Something that happens or exist during an event.
    \item \textsc{artifact} Something \textit{concrete} which is the result of the work of someone (e.g.\ written\_by, made\_by).
    \item \textsc{origin} Something \textit{abstract} which is originated by something else (e.g.\ invented, idea, title\_obtained\_by).
    \item \textsc{topic} The topic or focus of something.
    \item  \textsc{opposite} Something that is physically or idealistically opposite, contrary, against or inverse of something else.
    \item \textsc{cause-effect} An event or object which leads to an effect.
    \item \textsc{win-defeat} Someone or something who has won or lost a competition, an award or a war (default is victory, in case of defeat it is specified in the `Explanation' field).
    \item \textsc{type-of} The type, property, feature or characteristic of something.
    \item \textsc{named} Two spans which refer to the same entity (e.g.\ nickname, acronym, second name or abbreviation of something or someone).
    \item \textsc{related-to} Two semantically connected entities which do not fall in any of the previous cases.
\end{itemize}

\section{Entity Alteration Samples}
\label{app:data-samples}

In Table~\ref{tab:entities} we report one sample for each entity alteration type that we perform in respect to the original entity annotations from CrossNER~\cite{crossNER} and CoNLL-2003~\cite{tjong-kim-sang-de-meulder-2003-introduction}.
In the first sample, we correct the entity type of $e_1$ from `conference' to `organisation'.
In the second sample, we extend $e_2$---which originally only contains an adjective---in order to include also the following noun. We do this because, following previous work on RE, our relation labels do not hold between adjectives only.
Last, in the third sample we add the annotation for marking `Squealer' as an entity.

\begin{table}[]
    \centering
    \resizebox{0.9\columnwidth}{!}{
    \begin{tabular}{r|l}
        \toprule
        \textbf{Parameter} & \textbf{Value} \\
        \midrule
        Encoder & \texttt{bert-base-cased} \\
        Classifier & 1-layer FFNN \\
        Loss & Cross Entropy\\
        Optimizer & Adam optimizer \\
        Learning rate & $2e^{-5}$ \\
        Batch size & 32 \\
        Seeds & 4012, 5096, 8878, 8857, 9908 \\
        \bottomrule
    \end{tabular}}
    \caption{\textbf{Hyperparameters Setting.} Model details for reproducibility of the baseline.}
    \label{tab:reproducibility}
\end{table}

\begin{table*}[]
\resizebox{\textwidth}{!}{
\centering
\begin{tabular}{l}
\toprule
\textbf{\textsc{Entity Type}} \\
\hline
\begin{dependency}
\begin{deptext}
\emph{Finally, every other year,} \& \emph{ELRA} \& \emph{organizes a major conference} \& \emph{LREC} \& \emph{, the} \& \emph{International Language Resources and Evaluation Conference}\&\emph{.} \\ \\
\& $e_1$: conference \& \& $e_2$: conference \& \& $e_3$: conference \& \\
\end{deptext}
\wordgroup[group style={fill=orange!40, draw=brown}]{1}{2}{2}{entity1}
\wordgroup[group style={fill=orange!40, draw=brown}]{1}{4}{4}{entity2}
\wordgroup[group style={fill=orange!40, draw=brown}]{1}{6}{6}{entity3}
\end{dependency}\\
\begin{dependency}
\begin{deptext}
\emph{Finally, every other year,} \& \emph{ELRA} \& \emph{organizes a major conference} \& \emph{LREC} \& \emph{, the} \& \emph{International Language Resources and Evaluation Conference}\&\emph{.} \\ \\
\& $e_1$: organisation \& \& $e_2$: conference \& \& $e_3$: conference \& \\
\end{deptext}
\wordgroup[group style={fill=green!85!blue!30!white, draw=green!60!black}]{1}{2}{2}{entity1}
\wordgroup[group style={fill=orange!40, draw=brown}]{1}{4}{4}{entity2}
\wordgroup[group style={fill=orange!40, draw=brown}]{1}{6}{6}{entity3}
\end{dependency}\\

\midrule
\textbf{\textsc{Entity Boundaries}} \\
\hline
\begin{dependency}
\begin{deptext}
\emph{China} \& \emph{controlled most of the match and saw several chances missed until the 78th minute when} \& \emph{Uzbek} \& \emph{striker} \& \emph{Igor Shkvyrin} \& \emph{took advantage [...]} \\ \\
$e_1$: country \& \& $e_2$: misc \& \& $e_3$: person \& \\
\end{deptext}
\wordgroup[group style={fill=green!85!blue!30!white, draw=green!60!black}]{1}{1}{1}{entity1}
\wordgroup[group style={fill=orange!40, draw=brown}]{1}{3}{3}{entity2}
\wordgroup[group style={fill=yellow!40, draw=brown!50!yellow}]{1}{5}{5}{entity3}
\end{dependency}\\
\begin{dependency}
\begin{deptext}
\emph{China} \& \emph{controlled most of the match and saw several chances missed until the 78th minute when} \& \emph{Uzbek striker} \& \emph{Igor Shkvyrin} \& \emph{took advantage [...]} \\ \\
$e_1$: country \& \& $e_2$: misc \& $e_3$: person \& \\
\end{deptext}
\wordgroup[group style={fill=green!85!blue!30!white, draw=green!60!black}]{1}{1}{1}{entity1}
\wordgroup[group style={fill=orange!40, draw=brown}]{1}{3}{3}{entity2}
\wordgroup[group style={fill=yellow!40, draw=brown!50!yellow}]{1}{4}{4}{entity3}
\end{dependency}\\

\midrule
\textbf{\textsc{New Entities}} \\
\hline
\begin{dependency}
\begin{deptext}
\emph{Tamsin Greig} \& \emph{narrated, and the cast included} \& \emph{Nicky Henson} \& \emph{as} \& \emph{Napoleon} \& \emph{,}  \& \emph{Toby Jones} \& \emph{as the propagandist Squealer, and} \& \emph{Ralph Ineson} \& \emph{as} \& \emph{Boxer} \& \emph{.} \\ \\
$e_1$: person \& \& $e_2$: person \& \& $e_3$: person \& \& $e_4$: person \& \& $e_5$: person \& \& $e_6$: person \& \\
\end{deptext}
\wordgroup[group style={fill=blue!85!yellow!30!white, draw=blue!60!black}]{1}{1}{1}{entity1}
\wordgroup[group style={fill=blue!85!yellow!30!white, draw=blue!60!black}]{1}{3}{3}{entity2}
\wordgroup[group style={fill=blue!85!yellow!30!white, draw=blue!60!black}]{1}{5}{5}{entity3}
\wordgroup[group style={fill=blue!85!yellow!30!white, draw=blue!60!black}]{1}{7}{7}{entity4}
\wordgroup[group style={fill=blue!85!yellow!30!white, draw=blue!60!black}]{1}{9}{9}{entity5}
\wordgroup[group style={fill=blue!85!yellow!30!white, draw=blue!60!black}]{1}{11}{11}{entity6}
\end{dependency}\\
\begin{dependency}
\begin{deptext}
\emph{Tamsin Greig} \& \emph{narrated, and the cast included} \& \emph{Nicky Henson} \& \emph{as} \& \emph{Napoleon} \& \emph{,}  \& \emph{Toby Jones} \& \emph{as the propagandist} \& \emph{Squealer} \& \emph{, and} \& \emph{Ralph Ineson} \& \emph{as} \& \emph{Boxer} \& \emph{.} \\ \\
$e_1$: person \& \& $e_2$: person \& \& $e_3$: person \& \& $e_4$: person \& \& $e_A$: person \& \& $e_5$: person \& \& $e_6$: person \& \\
\end{deptext}
\wordgroup[group style={fill=blue!85!yellow!30!white, draw=blue!60!black}]{1}{1}{1}{entity1}
\wordgroup[group style={fill=blue!85!yellow!30!white, draw=blue!60!black}]{1}{3}{3}{entity2}
\wordgroup[group style={fill=blue!85!yellow!30!white, draw=blue!60!black}]{1}{5}{5}{entity3}
\wordgroup[group style={fill=blue!85!yellow!30!white, draw=blue!60!black}]{1}{7}{7}{entity4}
\wordgroup[group style={fill=blue!85!yellow!30!white, draw=blue!60!black}]{1}{9}{9}{entity5}
\wordgroup[group style={fill=blue!85!yellow!30!white, draw=blue!60!black}]{1}{11}{11}{entity6}
\wordgroup[group style={fill=blue!85!yellow!30!white, draw=blue!60!black}]{1}{13}{13}{entity6}
\end{dependency}\\
\bottomrule
\end{tabular}
}
\caption{\label{tab:entities}
\textbf{Samples of Modified Entity Annotations.} Instances with the original annotations from CrossNER~\cite{crossNER} and corresponding sentences from \projname{} with the corrected entities.
}
\end{table*}

\begin{figure}
\centering
\includegraphics[width=\columnwidth]{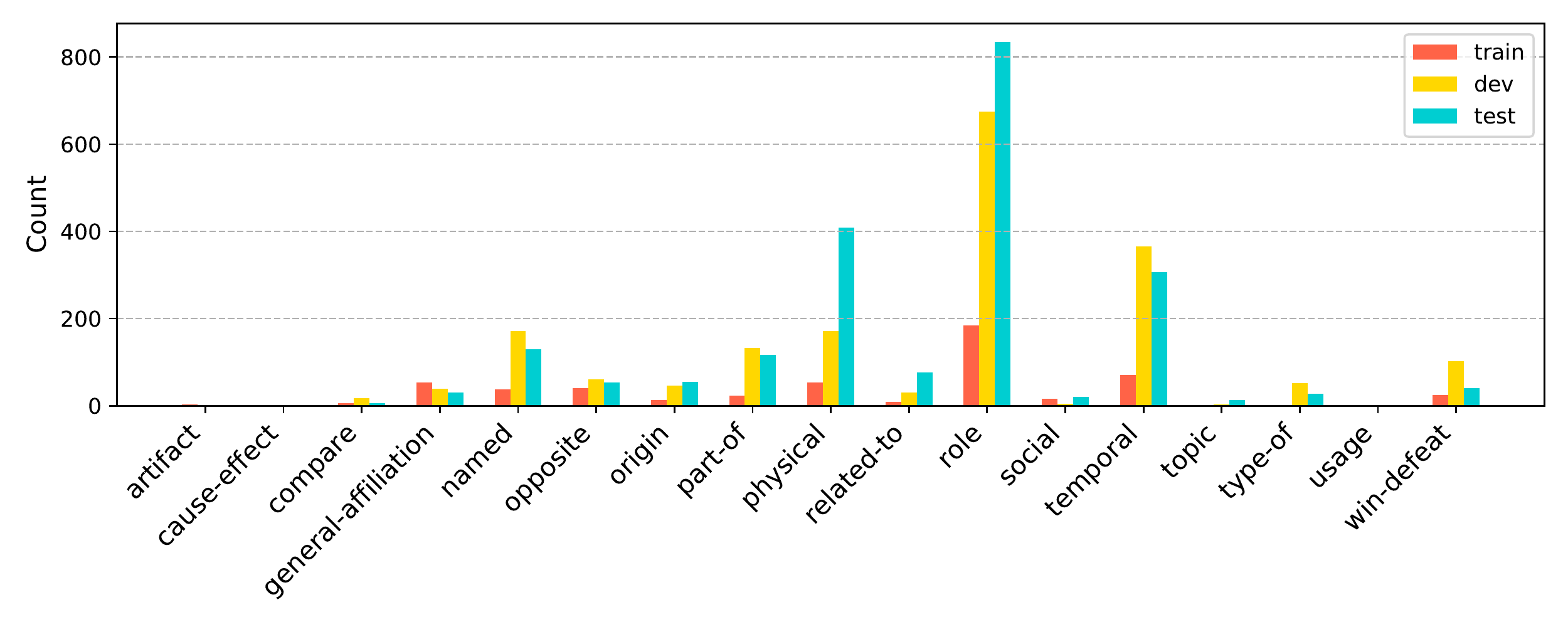}
\caption{\textbf{Label Distribution of the Politics Domain.} Distribution of the 17 relation types over the train/dev/test split.}
\label{fig:politics-label-distribution}
\end{figure}

\section{Detailed Label Statistics}
\label{app:statistics}

Table~\ref{tab:label-distribution} contains the detailed label statistics (counts and percentages) for each domain.

\begin{table*}[]
    \centering
    \resizebox{\textwidth}{!}{
    \begin{tabular}{r|r|l|r|l|r|l|r|l|r|l|r|l}
    \toprule
         & \multicolumn{2}{c|}{\textbf{News}} & \multicolumn{2}{c|}{\textbf{Politics}} & \multicolumn{2}{c|}{\textbf{Nat. Science}} & \multicolumn{2}{c|}{\textbf{Music}} & \multicolumn{2}{c|}{\textbf{Literature}} & \multicolumn{2}{c}{\textbf{AI}} \\
    \cmidrule(r){2-13}
        & Count & \% & Count & \% & Count & \% & Count & \% & Count & \% & Count & \% \\
    \midrule
    \textsc{artifact} & 1 & 0.11 & 6 & 0.13 & 70 & 2.16 & 612 & 12.48 & 620 & 17.26 & 99 & 3.8 \\
    \textsc{cause-effect} & 0 & 0.0 & 1 & 0.02 & 106 & 3.28 & 4 & 0.08 & 7 & 0.19 & 9 & 0.35 \\
    \textsc{compare} & 0 & 0.0 & 29  & 0.64 & 90 & 2.78 & 7 & 0.14 & 13 & 0.36 & 76 & 2.92 \\
    \textsc{general-affiliation} & 13 & 1.47 & 123 & 2.71 & 133 & 4.11 & 1,349 & 27.5 & 642 & 17.87 & 104 & 3.99 \\
    \textsc{named} & 32 & 3.62 & 338 & 7.45 & 251 & 7.76 & 164 & 3.34 & 209 & 5.82 & 306 & 11.74 \\
    \textsc{opposite} & 106 & 11.98 & 154 & 3.4 & 28 & 0.87 & 21 & 0.43 & 32 & 0.89 & 21 & 0.81 \\
    \textsc{origin} & 1 & 0.11 & 114 & 2.51 & 270 & 8.35 & 71 & 1.45 & 114 & 3.17 & 178 & 6.83 \\
    \textsc{part-of} & 47 & 5.31 & 273 & 6.02 & 246 & 7.61 & 421 & 8.58 & 243 & 6.76 & 517 & 19.83 \\
    \textsc{physical} & 276 & 31.19 & 634 & 12.98 & 481 & 14.87 & 782 & 15.93 & 348 & 9.69 & 259 & 9.93 \\
    \textsc{related-to} & 27 & 3.05 & 116 & 2.56 & 270 & 8.35 & 62 & 1.26 & 173 & 4.81 & 254 & 9.75 \\
    \textsc{role} & 275 & 31.07 & 1,695 & 37.38 & 716 & 22.14 & 767 & 15.64 & 578 & 16.09 & 269 & 10.32 \\
    \textsc{social} & 3 & 0.34 & 42 & 0.93 & 33 & 1.02 & 27 & 0.55 & 97 & 2.7 & 2 & 0.08 \\
    \textsc{temporal} & 2 & 0.23 & 744 & 16.41 & 41 & 1.27 & 78 & 1.59 & 117 & 3.26 & 65 & 2.49 \\
    \textsc{topic} & 3 & 0.34 & 17 & 0.37 & 95 & 2.94 & 13 & 0.27 & 97 & 2.7 & 88 & 3.38 \\
    \textsc{type-of} & 3 & 0.34 & 80 & 1.76 & 249 & 7.7 & 214 & 4.36 & 42 & 1.17 & 130 & 4.99 \\
    \textsc{usage} & 1 & 0.11 & 1 & 0.02 & 55 & 1.7 & 95 & 1.94 & 25 & 0.7 & 199 & 7.63 \\
    \textsc{win-defeat} & 95 & 10.73 & 167 & 3.68 & 100 & 3.09 & 222 & 4.53 & 236 & 6.57 & 30 & 1.15 \\
    \midrule
    total & 885 & & 4,534 & & 3,234 & & 4,909 & & 3,593 & & 2,606 \\
    \bottomrule
    \end{tabular}}
    \caption{\textbf{Relation Label Statistics.} Absolute count and relative percentage of each relation label. Note that, because of the multi-label setup, these numbers are higher in respect to the relation counts in Table~\ref{tab:statistics}.}
    \label{tab:label-distribution}
\end{table*}

\section{Multi-label annotation}
\label{app:multi-label-sample}

In Table~\ref{tab:multi-label-sample} we report an example of multi-label annotation in which $e_1$, a politician entity, is related to $e_3$, an election. The entity pair is annotated both as \textsc{temporal} because it provides temporal information about \emph{Tony Abbott}'s existence, and also as \textsc{win-defeat}, to capture the fact that he lost the election mentioned in $e_3$.

\begin{table*}[]
\resizebox{\textwidth}{!}{
\centering
\begin{tabular}{l}
\toprule
\textbf{\textsc{Multi-label Annotation}} \\
\hline
\begin{dependency}
\begin{deptext}
\emph{He was the last former Prime Minister to lose his seat until} \& \emph{Tony Abbott} \& \emph{lost his seat of} \& \emph{Warringah} \& \emph{in} \& \emph{2019 Australian federal election} \& \emph{, [...]} \\ \\
\& $e_1$: politician \& \& $e_2$: location \& \& $e_3$: election \& \\
\end{deptext}
\wordgroup[group style={fill=orange!40, draw=brown}]{1}{2}{2}{entity1}
\wordgroup[group style={fill=yellow!40, draw=brown!50!yellow}]{1}{4}{4}{entity2}
\wordgroup[group style={fill=blue!85!yellow!30!white, draw=blue!60!black}]{1}{6}{6}{entity3}
\end{dependency}\\
($e_1$, $e_3$, \textsc{temporal}) ($e_1$, $e_3$, \textsc{win-defeat}) \\
\bottomrule
\end{tabular}
}
\caption{\label{tab:multi-label-sample}
\textbf{Example of Multi-label Annotation.} Example from \projname{} of an ordered entity pair which has been annotated with multiple relation labels.
}
\end{table*}

\section{Reproducibility}
\label{app:reproducibility}

We report in Table~\ref{tab:reproducibility} the hyperparameter setting of our RC model (see Section~\ref{sec:model}).
All experiments were ran on an NVIDIA\textsuperscript{®} A100 SXM4 40 GB GPU and an AMD EPYC\textsuperscript{™} 7662 64-Core CPU.

\section{Label Distribution Per-Domain}
\label{app:politics-distribuition}

Given the imbalance of the label distribution 
(see Figure~\ref{fig:label-distribution}) and of the train/dev/test splits (see Table~\ref{tab:statistics}), we report in Figure~\ref{fig:politics-label-distribution} as a sample the specific label distribution of the politics domain.

\end{document}